\documentclass[manuscript,screen]{acmart}
\AtBeginDocument{%
  \providecommand\BibTeX{{%
    \normalfont B\kern-0.5em{\scshape i\kern-0.25em b}\kern-0.8em\TeX}}}

\setcopyright{acmcopyright}
\copyrightyear{2018}
\acmYear{2018}
\acmDOI{XXXXXXX.XXXXXXX}

\acmConference {LAK2024}{March,18-22,
  2024}{Kyoto, Japan}
\acmBooktitle{Woodstock '18: ACM Symposium on Neural Gaze Detection,
 June 03--05, 2018, Woodstock, NY} 
\acmPrice{15.00}
\acmISBN{978-1-4503-XXXX-X/18/06}

\begin{document}

\title{Kattis vs. ChatGPT: Assessment and Evaluation of Programming Tasks in the Age of Artificial Intelligence }

\author{Dunder, N.}
\email{trovato@corporation.com}
\author{Lundborg, S.}
\authornotemark[1]
\email{webmaster@marysville-ohio.com}
\affiliation{%
  \institution{KTH Royal Institute of Technology}
  \streetaddress{P.O. Box 1212}
  \city{Stockholm}
  \state{Stockholm}
  \country{Sweden}
  \postcode{43017-6221}
}

\author{Wong, J.}
\affiliation{%
  \institution{Utrecht University}
  \streetaddress{}
  \city{Hekla}
  \country{Netherlands}}
\email{larst@affiliation.org}

\author{Viberg, O.}
\affiliation{%
  \institution{KTH Royal Institute of Technology}
  \city{XX}
  \country{Sweden}
}

\begin{abstract}
AI-powered education technologies can support students and teachers in computer science education. However, with the recent developments in generative AI, and especially the increasingly emerging popularity of ChatGPT, the effectiveness of using large language models for solving programming tasks has been underexplored. The present study examines ChatGPT's ability to generate code solutions at different difficulty levels for introductory programming courses. We conducted an experiment where ChatGPT was tested on 127 randomly selected programming problems provided by \textit{Kattis}, an automatic software grading tool for computer science programs, often used in higher education. The results showed that ChatGPT independently could solve 19 out of 127 programming tasks generated and assessed by \textit{Kattis}. Further, ChatGPT was found to be able to generate accurate code solutions for simple problems but encountered difficulties with more complex programming tasks. The results contribute to the ongoing debate on the utility of AI-powered tools in programming education.
\end{abstract}


\maketitle

\section{Introduction}
Programming education is "an important source of skills and knowledge for students and a necessary feature to survive in a competitive job market" (~\cite{yilmaz2023effect}, p.2). During the last decade, artificial intelligence(AI)-powered education technologies have been used in programming education to assist teachers in guiding students to acquire programming and computational skills and knowledge~\cite{geng2023can}. Such AI-powered technologies include various types of intelligent tutoring systems (for an overview, see~\cite{crow2018intelligent}, plagiarism detection tools ~\cite{cheers2021academic} and auto-grading tools such as \textit{Kattis} (e.g.,~\cite{basnet2018exploring})). Therefore, AI is not futuristic but actively used in classrooms and courses worldwide~\cite{gavsevic2023empowering}. 

Recently, new kinds of AI-powered tools, namely generative AI technologies - "a distinct class of AI and an incredibly powerful technology that has been popularized by ChatGPT" (~\cite{lim2023generative},p.2) - have rapidly penetrated different parts of our society, including learning and teaching practices in the setting of programming higher education ~\cite{geng2023can,yilmaz2023effect, tian2023chatgpt, kiesler2023large}. Whereas earlier research on using AI-powered education tools in programming education has shown related evidence in terms of their efficacy to support students in coding tasks by, for example, providing suggestions~\cite{sharma2022designing}, error detection~\cite{huang2019ai}, and automatic code generation ~\cite{basnet2018exploring}, our understanding of ChatGPT and other rapidly emerging large language models' effectiveness for accurately solving coding tasks is still limited~\cite{kiesler2023large,tian2023chatgpt}. Knowing the capabilities of these tools is important since they are already extensively used by students and can generate the code from description, perform code completion, translation, and summarization ~\cite{lu2021codexglue}. 

Evidence for the power and potential to solve programming tasks using ChatGPT in introductory-level programming education is emerging ~\cite{geng2023can}. All this offers several new opportunities to improve students' conditions of learning. However, at the same time, it raises several questions and concerns, including those that pertain to the degree of accuracy of the provided solutions and the student's intention to use such tools to complete course assignments. Concerns were raised by ~\cite{becker2023programming}, given that it has become much easier for students to produce codes to pass traditional first-year programming assignments and even exams using generative AI-powered tools. This brings us to a heavily debated concern about academic integrity in higher education in the era of generative AI and, on the other hand, the opportunities for enhancing learning and teaching ~\cite{sullivan2023chatgpt}. 

Central to these concerns is a larger question about ChatGPT's capacity to empower students in programming education. As stressed by Steele ~\cite{steele2023gpt}, AI chatbots such as ChatGPT can threaten contemporary education systems in terms of: "(1) measurement of students’ knowledge and skills; (2) accuracy of the information students are learning; and (3) the market value of the skills we are teaching" (p.1). Addressing such concerns requires examining the kinds of tasks AI chatbots can perform, and how well they can perform. Answers to these questions will offer us an evidence-based ground to consider the degree to which ChatGPT could be effectively integrated into programming education to improve students' conditions for learning.

The present study aims to evaluate the extent to which ChatGPT-3.5 can accurately solve programming tasks (at varied difficulty levels) provided and corrected by an automated code generation and assessment tool, \textit{Kattis}, which is widely used in introductory programming courses of engineering education at several highly ranked higher educational institutions across countries. The focus on the introductory level of programming education, in which the present study has been conducted, is important since, at this level, students need extensive practice in writing codes, among other activities, to gain fundamental programming skills ~\cite{towell2010walls}. Our study contributes to building a better understanding of ChatGPT's capabilities to inform teaching and learning practices by answering our main research question: \textit{To what extent is ChatGPT able to solve automatically generated coding tasks in the setting of introductory programming education?}


\section{Background}

\subsection{Automatic assessment of programming tasks}


The development of coding assignments and their assessment in programming education is a highly time-consuming task for educators ~\cite{fernandez2023automatic}, especially in the context of large courses (i.e.,>100 students participate in one course), as in the case of the present study. During the last few years, and especially from 2021, such task has been supported by AI-driven code generation tools such as \textit{OpenAI Codex} and \textit{Amazon CodeWhispee}\textit{r}, which are argued to be able to support students and educators in their everyday educational practices ~\cite{becker2023programming, kazemitabaar2023studying}. Such tools can be used to support educators by automatically generating programming exercises at various levels of difficulty and by generating guiding hints to coding solutions, like in the case of \textit{Kattis}. OpenAI and DeepMind have recently introduced groundbreaking generative AI-models such as ChatGPT-3.5 that are capable of not only generating coding assignments but also - computer code, potentially making programming more productive and accessible ~\cite{becker2023programming}. Such tools have hitherto been freely accessible to students, suggesting that some students participating in programming education are using AI code compilation in their coding assignments.

In the current study, we examine ChatGPT-3.5's solutions to programming tasks provided and assessed by \textit{Kattis}. \textit{Kattis}, available freely on (open.kattis.com), is an automated code generation and grading system introduced in programming courses at a large European university in 2002 and has since gained widespread popularity in higher education. Its primary purpose is not only to generate programming tasks automatically but also to shift the responsibility of assessing the correctness of program code from the instructor to an automated tool, thereby releasing the instructor's time that can instead be used to assist students continuously during lab sessions and other related tasks. Additionally, \textit{Kattis} serves as an online judge for submissions in programming competitions~\cite{enstrom2010testdriven}. 


\subsection{GhatGPT in higher education}
ChatGPT is most popularly used in higher education settings compared to K-12 education and training of practical skills ~\cite{hadi2023exploring}. As stressed by ~\cite{dai2023reconceptualizing}, students' self-initiated adoption of ChatGPT has made it almost impossible to ban or control it. Its rapid uptake rate among students has induced a student-driven educational tool" (p.86). The release and rapid diffusion of ChatGPT have caught educators' attention worldwide due to its and other AI-based technologies to "transform education" (~\cite{futterer2023chatgpt},p.1). 

On the one hand, early research exploring teachers' attitudes toward using ChatGPT has shown that educators are generally cautious in their approach to using ChatGPT (~\cite{iqbal2022exploring}). On the other hand, the results of recent studies have shown that students – another key stakeholder of generative AI tools in education – overall demonstrate positive attitudes toward ChatGPT but raised concerns about privacy, ethical issues, the impact on personal development and career prospects~\cite{chan2023students}.
A recent review  ~\cite{albadarin2023systematic} of 14 empirical studies showed that ChatGPT was beneficial for learning and teaching. Concerning learning, the reviewed studies showed that students used ChatGPT in a various ways: as an intelligent assistant for answering on-demand questions, searching for information, and receiving feedback. However, the findings on students' perceived accuracy, relevance, and reliability of ChatGPT's output are mixed. 

Among early results summarizing the use of ChatGPT in higher education (N = 12 papers examined), scholars highlight that the implementation of ChatGPT in education has a positive influence on the teaching and learning process but also stress the importance of teachers being trained to use the tool effectively~\cite{montenegro2023impact}. In programming education, teachers and students have similarly indicated that generative AI-powered tools would play a significant role. They also stressed several concerns about how large language models should be best integrated to support their needs ~\cite{zastudil2023generative}. The examination of ChatGPT's performance against AI-powered EdTech tools used in higher education for some time is so far limited. While some evidence on ChatGPT is emerging, it is still in its infancy due to the lack of rigorous evaluations of the impact of the use of ChatGPT on learning and teaching. Evaluating ChatGPT's capabilities, for example, in solving programming tasks, is needed to make well-informed teaching, assessment, and evaluation decisions.

\subsection {ChatGPT in programming education}
Programming skills and knowledge are becoming increasingly important in today's world that is rapidly evolving with the acceleration of technological and digital advancement(~\cite{yilmaz2023effect}. This study, in particular, focuses on \textit{introductory }programming education since it is considered to be especially challenging for students due to their inability to comprehend what is happening to their program in memory because they are incapable of creating a clear mental model of its execution ~\cite{milne2002difficulties}. With the emergence of ChatGPT in programming education, scholars argue that "[g]enerative AI-powered tools can transform programming education" (~\cite{yilmaz2023effect}, p.2) by using it as a bot for discussing source code and even generating code ~\cite{tian2023chatgpt}, among others.  

Whereas ChatGPT could be used in several ways in programming education, 
little is known about its performance in solving programming problems in introductory programming education. Geng and colleagues ~\cite{geng2023can} have explored how well ChatGPT (treated as one of the students) can perform in an introductory-level functional language programming course and found that it can achieve a grade \textit{B}, thus successfully passing the course. At the same time, the authors stress the importance of using ChatGPT in tandem with other teaching methods to ensure that students develop a well-rounded set of programming skills ~\cite{geng2023can}. Another recent study investigated the performance of ChatGPT-3.5 and GPT-4 in solving programming tasks ~\cite{kiesler2023large}. The results of evaluating 72 Python tasks' solutions (retrieved from the open source platform \textit{CodingBat}) for novice programmers were compared to the ChatGPT's performance. The findings demonstrated high scores of 94.4 to 95.8 percent correct responses. However, the authors stress that model solutions to all \textit{CodingBat} tasks are available in GitHub, suggesting that the chances that ChatGPT was trained in such data are high. Finally,~\cite{savelka2023can} evaluated the capability of ChatGPT to pass assessments in introductory and intermediate Python programming courses. The results showed that the current models are not capable of passing the full spectrum of assessments included in a Python programming course (<\%70 on even entry-level modules). All in all, the related evidence is scarce and inconsistent, suggesting that more empirical research is needed.

\section{Method}
\subsection{Study Design}
The present study was conducted in the setting of a large introductory programming course (i.e., > 100 students) that is a mandatory part of several engineering programs at a large technical university in Europe. \textit{Kattis}, an automated code-generation and grading tool has been used in the targeted setting of programming education for several years, and was found to be efficient in supporting teachers and students ~\cite{basnet2018exploring}. 
Coding tasks from \textit{Kattis} were sent to Open AI's freely accessible large language model, ChaptGPT-3.5, to generate solutions in Python code.
The assessment results generated by \textit{Kattis} were considered as the dependent variable, and the solutions provided by ChatGPT as the independent variable.  

\subsection{Coding Tasks}
The coding tasks (N = 127) were randomly selected from \textit{Kattis} in spring 2023,  with consideration for an even distribution across difficulty levels. Figure 1 illustrates the distribution of coding tasks across several difficulty levels. Each task in \textit{Kattis} comes with a metadata, indicating the difficulty level of the task, ranging from 1 to 10, and further categorized as 'easy' (1.0-2.7), 'medium' (2.8-5.3), and 'hard' (5.4-10.0). The difficulty level of a coding task is estimated by using a variant of the Elo rating system (see e.g.,  ~\cite{pelanek2016applications}). Specifically, whereas the tasks that have been solved by many people with only few attempts indicate a lower degree of difficulty, the tasks that have been attempted to be solved by many individuals but rarely solved indicate a higher level of difficulty. Tasks that have very few submissions tend to fall under the category of 'medium' difficulty as \textit{Kattis} lacks sufficient data on their difficulty level (https://open.kattis.com/help/ranklist). In this study, difficulty levels of the coding tasks were re-categorized as integers from 1 to 10 for clearer result visualization. That is, all tasks with a difficulty level between 1.0 to 1.9 are classified as tasks of difficulty Level 1. Further, those tasks that correspond to difficulty Levels 1 and 2 are grouped within a larger category of 'easy. Difficulty levels of 3 and 4 are grouped as 'medium', and difficulty levels from 5 to 10 were grouped as 'hard'. 
All the coding tasks were copied directly from \textit{Kattis}. One constraint relates to the fact that the choice of the coding tasks was limited to those that were possible to copy and transfer to ChatGPT easily; the tasks with a large number of mathematical equations and figures (common to task with high levels of difficulty) were excluded due to the technical limitations. 

Many of the coding tasks require solutions consisting of code ranging from 3 to 20 lines, with an average of around 10 lines. Most of the tasks can be solved by correctly formatting the input data and then performing a task. In many tasks, data is formatted into lists and dictionaries, sometimes tuples or sets. The operation performed is often either a for loop, a while loop, a mathematical operation, or sorting, possibly followed by an 'if-else' statement where the result is printed.

\begin{figure}[h]
  \centering
  \includegraphics[width=0.7\linewidth]{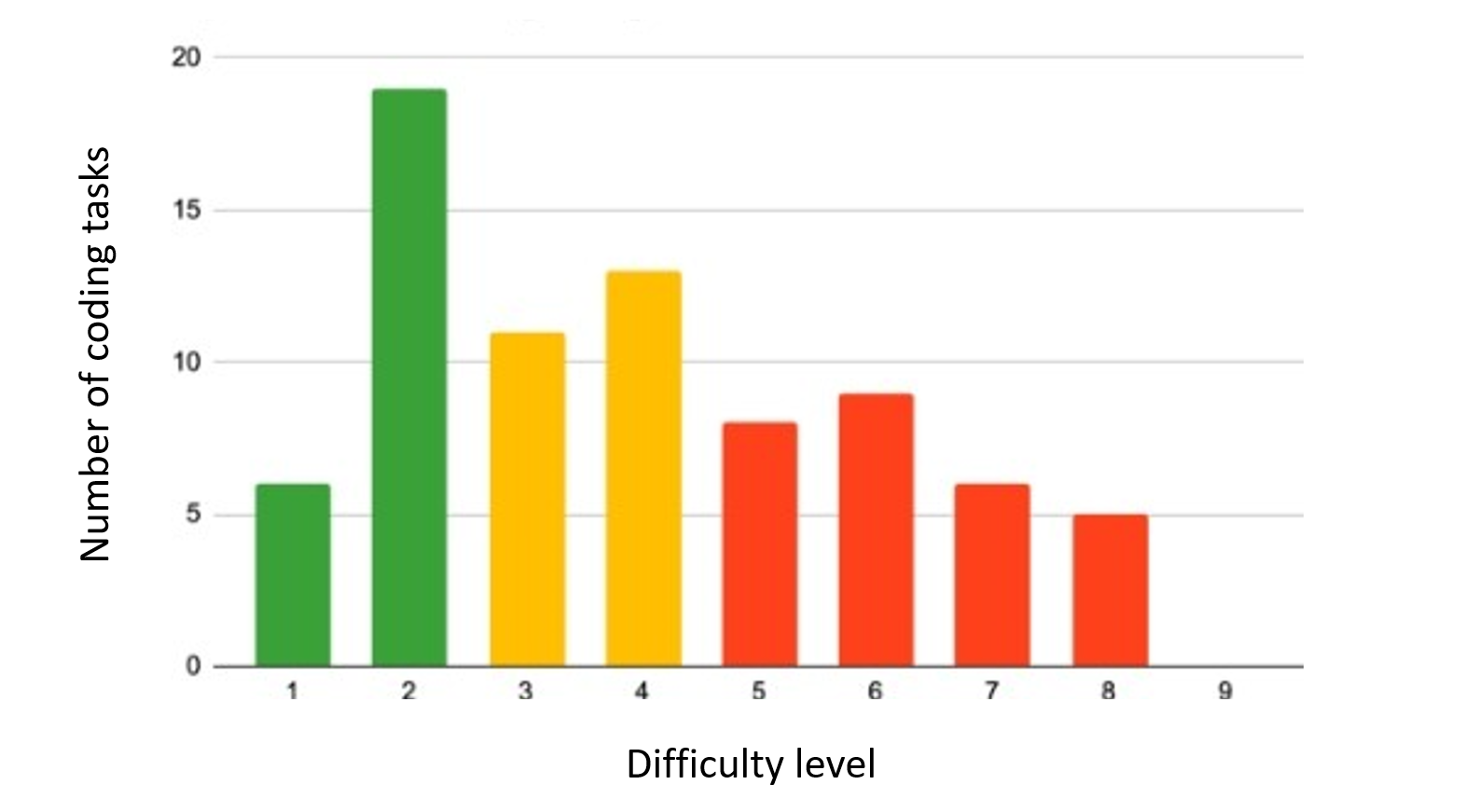}
  \caption{Distribution of the number of coding tasks across the difficulty levels}
  \Description{from the results table}
\end{figure}

Apart from the difficulty level, each coding task in \textit{Kattis }comes with a metadata. Table 1 provides a description of the metadata that was retrieved from \textit{Kattis }to examine the association with the assessment of the coding solutions provided by ChatGPT.

\begin{table*}
    \caption{Description on metadata retrieved from Kattis}
  \begin{tabular}{p{3cm}p{6cm}}  
    \toprule
   \textbf{Metadata} & \textbf{Description} \\
    \midrule
    \textit{Approved Submissions} & \% approved submissions of all submitted solutions in Kattis. Interval between 0-100. \\
    \hline
    \textit{Successful submitters} & \% of users who tried and then succeeded in solving the code problem. Range between 1-100. \\
    \hline
    \textit{Level of difficulty} & A number on a scale of 1-10, where 1 is the easiest and 10 the most difficult. \\
    \bottomrule
  \end{tabular}
\end{table*}

\subsection{Study Procedure}
Each coding task from \textit{Kattis} was sent to ChatGPT to generate a solution proposal in Python code. The proposed solution was then submitted to \textit{Kattis'} own text window, which returned a response as to whether the solution is approved or not. The type of error message from \textit{Kattis} (i.e., feedback) was noted in cases where an approved verdict could not be given. 

\subsection{Data Analysis}
To assess the performance of ChatGPT, we first determined the percentage of problems solved (i.e., solutions approved by \textit{Kattis}). We have also examined ChatGPT's performance for the level of the task's difficulty, where there was at least one approved solution. For the solutions that were not approved by \textit{Kattis}, the frequency and type of error messages were analyzed. To further examine the extent of ChatGPT's performance against the general performance on the coding tasks, we finally performed a correlational analysis on the number of approved solutions from ChatGPT in \textit{Kattis} (i.e., coded as a binary variable where '1' is accepted and '0' is not accepted) and the metadata retrieved from \textit{Kattis} (Table 1).

\section{Results}
\subsection{Performance of ChatGPT- approved solutions}
Out of the 127 code solutions generated by ChatGPT, only 19(15\%) were fully approved by \textit {Kattis}. Figure 2 illustrates the overall distribution of the coding tasks and the difficulty level. Green bars represent approved solutions and blue bars are the solutions that were not approved by \textit{Kattis}. Among the 19 approved solutions, ten were solutions to the coding tasks with the difficulty Level 1 (8\% out of overall sample of 127 tasks), seven corresponded to the difficulty Level 2 (6\%), and only two - the difficulty Level 4 (2\%). The majority (85\%) of the coding solutions generated by ChatGPT were not approved by\textit{ Kattis}. The results suggest a low performance by ChatGPT in terms of both correctly solving the tasks and its ability to solve high-level difficulty tasks.  

Further inspection shows that, among all the approved solutions, the approved solution for the coding task with the lowest level of difficulty is for task with a difficulty level of 1.3, while the one with the highest level of difficulty was rated at 4.2. Given that the task description, provided by \textit{Kattis }for the easiest task is much shorter than the task description for the most challenging task, ChatGPT also provided solutions of different length and quality: easiest task consists of only 11 lines of code and mainly 'if'- and 'else'-statements, while the most challenging task is composed of 55 lines of code and include 'loops', 'functions', 'lists', and 'matrices'. 


\subsection{Performance of ChatGPT- failed solutions}
\textit{Kattis} provided feedback in the form of error messages for the 108 coding tasks with incorrect coding solutions (i.e., ChatGPT solutions that were not approved by \textit{Kattis}). The \textit{'Wrong Answer'} feedback occurred for 83 tasks (77\% of all incorrect solutions) where \textit{Kattis} did not approve the code solution. The second most common error was \textit{'Run Time Error'} that occurred in 16 tasks (15\%), and the least common one was \textit{'Time Limit Exceeded'} that occurred for 9 tasks (8\%). This suggests that program crashes were the most frequent issue while running for too long was the least common problem.  

Out of the 108 incorrect coding solutions, 19 had indicated partially accepted solutions, referring to the coding solutions that were accepted for certain inputs and not all the inputs. This accounts for 18\% of failed solutions and 15\% of all solutions. For each coding task,\textit{Kattis} assessed the solution for a different number of inputs. Table 2 provides an overview of the ratio of inputs accepted by \textit{Kattis}, the assessment (i.e., the error message shown to the submitter as feedback on the coding solution), and difficulty level of the partially accepted coding tasks. 
In this set of coding tasks, \textit{'Wrong Answer'} occurred for most of the coding tasks, followed by textit{'Time Limit Exceeded'} and \textit{'Run Time Error'}. 

Among the set of partially accepted solutions, the most number of inputs that were approved (i.e., n = 12) was found for two tasks, one with difficulty level 3.9 and the other with difficulty level 4.1. The coding solution that had the highest percentage of inputs accepted was of difficulty level 2.8; it was accepted for 60\% of the inputs. The coding solution for the task with the highest difficulty level (i.e., 6.5) was accepted for only 1 out of 13 inputs. This suggests a varying performance of ChatGPT across tasks of different difficulty levels.



\begin{figure}
  \centering
  \includegraphics[width=0.7\linewidth]{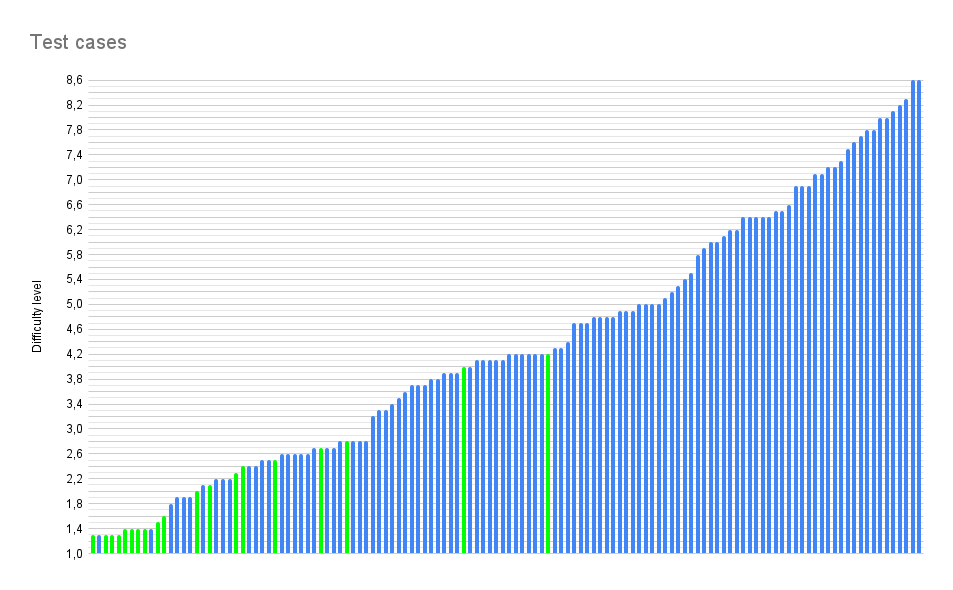} 
  \caption{Distribution graph of the 127 tasks at different difficulty levels (1-10). Green bars represent tasks approved by \textit{Kattis}, and blue bars – not approved.}
\end{figure}

\begin{table*}
  \caption{Overview of the type of errors and difficulty level of partially accepted solutions}
  \begin{tabular}{p{2cm}p{2cm}p{2cm}p{2cm}}  
    \toprule
    Ratio of approved inputs & \% of approval & Kattis' assessment & Difficulty level \\
    \midrule
    1/13 & 8\% & WA & 1.9 \\
    3/12 & 25\% & TLE & 2.2 \\
    1/33 & 3\% & WA & 2.4 \\
    1/38 & 3\% & WA & 2.5 \\
    5/23 & 22\% & WA & 2.6 \\
    1/13 & 8\% & WA & 2.6 \\
    1/78 & 1\% & WA & 2.7 \\
    5/28 & 18\% & WA & 2.6 \\
    3/5 & 60\% & TLE & 2.8 \\
    1/34 & 3\% & TLE & 3.2 \\
    3/11 & 27\% & WA & 3.3 \\
    2/13 & 15\% & WA & 3.5 \\
    12/21 & 57\% & WA & 3.9 \\
    12/40 & 30\% & WA & 4.1 \\
    1/2 & 50\% & RTE & 4.8 \\
    2/52 & 4\% & WA & 5.0 \\
    5/18 & 28\% & TLE & 6.2 \\
    2/17 & 12\% & TLE & 6.4 \\
    1/13 & 8\% & WA & 6.5 \\
    \bottomrule
  \end{tabular}
\end{table*}

\subsection{Correlations analysis: Evaluation of ChatGPT's performance against general performance}
Table 3 shows the the correlational table for ChatGPT's performance in terms of the accepted solutions and the general performance indicators on the selected coding tasks, based on the metadata (for description of the metadata, see Table 1). Positive correlations of 0.43 and 0.37 were found between approved solutions of ChatGPT and the number of submissions for the coding tasks, and between approved solutions of ChatGPT and successful 'submitters' respectively. Therefore, ChatGPT's ability to solve a coding task correlates slightly more with the approved submissions than with the number of successful submitters. The negative correlation of -0.48 between the level of difficulty and approved solutions of ChatGPT indicates that as the difficulty level of the coding tasks increases, the number of approved solutions from ChatGPT decreases. Overall, the results suggest that ChatGPT performs better on easier coding tasks that has high percentage of submissions that were already approved in the system. 

\begin{table*}
      \caption{Correlations between ChatGPT's performance and indicators of general performance}
      \begin{tabular}{ccccccc}
        \toprule
    Variable & M & SD & 1 & 2 & 3 & 4 \\
        \midrule
    1. Approved Submissions & 36.8 & 13.5 & - & & & \\
    2. Successful Submitters & 81.8 & 12.4 & 0.71*** & - & & \\
    3. Difficulty Level & 4.3 & 2 & --0.70*** & --0.84*** & - & \\
    4. Approved Solutions of ChatGPT (=1) & 0.1 & 0.4 & 0.43*** & 0.37*** & --0.48*** & - \\
    \midrule
    * $P \leq 0.05$, ** $P \leq 0.01$, *** $P \leq 0.001$
      \end{tabular}
\end{table*}

\section{Discussion and conclusions}
Considering that generative AI-powered tools can transform programming education ~\cite{yilmaz2023effect}, many questions remain regarding how this transformation could be facilitated, ultimately leading to the student-improved acquisition of programming skills and knowledge. To fill this gap, this study examined the extent to which ChatGPT is capable of solving automatically generated coding tasks in the setting of introductory programming education. By examining ChatGPT's ability to solve coding tasks (at different levels of difficulty), automatically generated and assessed by\textit{ Kattis,} (i.e., the system that is frequently used in computer science education for the provision of feedback and assessment across countries), we found that the current capability of ChatGPT to solve such tasks in the targeted setting is somewhat limited (only 15\% were approved by \textit{Kattis}). These results are not supported by earlier research findings (e.g., ~\cite{geng2023can,kiesler2023large}), but are in line with the results of another recent study by ~\cite{savelka2023can}. Our results also give us a nuanced picture of the difficulty level of programming tasks that ChatGPT can solve, as shown by how it can currently solve mostly the tasks categorised as 'easy'. 

In general, these findings imply that over-reliance on the code solutions provided by ChatGPT today may hamper students' acquisition of programming skills and knowledge, and can be especially detrimental for students in introductory programming education. However, this does not suggest that ChatGPT should be banned since as shown by ~\cite{yilmaz2023effect}, its use, as experienced by students, can improve their thinking skills, increase their self-confidence when solving programming tasks, and to facilitate debugging. Instead, educators are recommended to design and integrate teaching activities that would enable students to critically reflect and use of ChatGPT for learning by self-evaluating the responses (to programming tasks) provided by generative AI-powered tools such as ChatGPT or similar. To achieve this, educators need to be supported in terms of their professional development focusing on teaching with AI, since many of them may lack relevant knowledge and skills that constitute AI digital competencies (e.g.,~\cite{ng2023teachers}). 

Furthermore, since ChatGPT's use has been predominantly driven by students ~\cite{dai2023reconceptualizing}, they need to be supported in the development of relevant skills, including their critical thinking- and self-regulated learning (SRL) skills, which are positively associated with their academic performance in online and blended learning settings~\cite{xu2023synthesizing}, and in which the prevalent part of introductory programming education is offered. Such support can be consider in several ways. One way is to ensure that the fostering of students' critical thinking- and SRL skills is a part of the introductory program education curriculum. Another complementary way is to explore the opportunities of how ChatGPT can be used to effectively support students in their development of such skills. As recently demonstrated by scholars, using AI applications for supporting students' SRL in online learning can be effective~\cite{jin2023supporting}. 

This study has several limitations. First, only 127 programming tasks have been tested. This limits our ability to generalise the results. Second, the tests were performed over one month on different days, and during that time, ChatGPT-3.5 was updated (March 23, 2023), which may have influenced our results. For more controlled experimental settings, we recommend performing all the tests on the same day. Third, the study assumes how students might generate code solutions from ChatGPT by copying and pasting the task descriptions provided by \textit{Kattis}. It is not known if students might use ChatGPT in other ways, such as using prompt engineering to generate partial codes and evaluating the accuracy of the codes themselves. 

To better understand the key stakeholders' perspectives on utilising ChatGPT in teaching and learning programming skills, future research should focus on following up qualitative research studies targeting both students and teachers. Second, since new language language models are continuously developing, scholars need to continuously evaluate their capabilities in authentic settings of introductory programming education. Third, studies in which ChatGPT is assisted by a human (either the teacher or the student) in solving tasks at higher difficulty levels are recommended to gain a deeper insight into the benefits and limitations of Human-AI collaboration for student-improved learning of programming skills. Finally, rigorous experimental studies that measure changes in student learning are needed to assess the impact of utilising ChatGPT or similar chatbots on student acquisition of programming skills and their academic performance.

In sum, our study provides insights into ChatGPT's performance and constraints when evaluated against the AI-powered EdTech code generation and auto-grading tools used in programming higher education for some time (i.e., \textit{Kattis}). The results show that ChatGPT performs well in solving only easy-level programming tasks, which indicates its limited capability to be used by students to pass introductory programming courses. The findings contribute to the ongoing debate on the utility of AI-powered tools in introductory programming education.

\bibliographystyle{ACM-Reference-Format}
\bibliography{mybibliography}

\end{document}